%% file: main.tex
\documentclass[conference]{IEEEtran}
\IEEEoverridecommandlockouts
\usepackage{cite}
\usepackage{url}
\usepackage{amsmath,amssymb,amsfonts}
\usepackage{algorithmic}
\usepackage{graphicx}
\usepackage{textcomp}
\usepackage{xcolor}
\usepackage{comment}
\usepackage{multirow}
\usepackage{mdframed}
\usepackage{tikz}
\def\BibTeX{{\rm B\kern-.05em{\sc i\kern-.025em b}\kern-.08em
    T\kern-.1667em\lower.7ex\hbox{E}\kern-.125emX}}

\usepackage{listings}
\definecolor{codegreen}{rgb}{0,0.6,0}
\definecolor{codegray}{rgb}{0.5,0.5,0.5}
\definecolor{codepurple}{rgb}{0.58,0,0.82}
\definecolor{backcolour}{rgb}{1.0,1.0,1.0}
{\renewcommand{\baselinestretch}{0.96}

\lstdefinestyle{std}{
    backgroundcolor=\color{backcolour},   
    commentstyle=\color{codegreen},
    keywordstyle=\color{magenta},
    numberstyle=\ttfamily\scriptsize\color{codegray},
    stringstyle=\color{codepurple},
    basicstyle=\ttfamily\scriptsize,
    breakatwhitespace=false,         
    breaklines=true,                 
    captionpos=b,                    
    keepspaces=true,                 
    numbers=left,                    
    numbersep=2pt,                  
    showspaces=false,                
    showstringspaces=false,
    showtabs=false,                  
    tabsize=2
}

\definecolor{backcolour_high}{rgb}{1.,0.935,0.602}
\lstdefinestyle{highlight}{
    backgroundcolor=\color{backcolour_high},   
    commentstyle=\color{codegreen},
    keywordstyle=\color{magenta},
    numberstyle=\ttfamily\scriptsize\color{codegray},
    stringstyle=\color{codepurple},
    basicstyle=\ttfamily\scriptsize,
    breakatwhitespace=false,         
    breaklines=true,                 
    captionpos=b,                    
    keepspaces=true,                 
    numbers=left,                    
    numbersep=2pt,                  
    showspaces=false,                
    showstringspaces=false,
    showtabs=false,                  
    tabsize=2
}

\lstnewenvironment{std}[1]
{\lstset{style=std,
         language=Python,
         firstnumber=#1}}
{}

\lstnewenvironment{highlight}[1]
{\lstset{style=highlight,
         language=Python,
         firstnumber=#1}}
{}

\lstnewenvironment{highlight_last}[3]
{\lstset{style=highlight,
         language=Python,
         firstnumber=#1,
         caption=#2,
         label=#3}}
{}

\newcommand{\rev}[1]{#1}

\newcommand\copyrighttext{%
  \footnotesize \textcopyright 2023 IEEE. Personal use of this material is permitted.  Permission from IEEE must be obtained for all other uses, in any current or future media, including reprinting/republishing this material for advertising or promotional purposes, creating new collective works, for resale or redistribution to servers or lists, or reuse of any copyrighted component of this work in other works.
 
  Accepted at the 2023 Forum on Specification \& Design Languages (FDL).}
\newcommand{\copyrightnotice}{%
\begin{tikzpicture}[remember picture,overlay,scale=1.00, every node/.style={scale=1.00}]
\node[anchor=south,yshift=10pt] at (current page.south) {\fbox{\parbox{\dimexpr\textwidth-\fboxsep-\fboxrule\relax}{\copyrighttext}}};
\end{tikzpicture}%
}

\begin{document}
\bstctlcite{IEEEexample:BSTcontrol}

\title{PLiNIO: A User-Friendly Library of Gradient-based Methods for \rev{Complexity}-aware DNN Optimization
% \thanks{Identify applicable funding agency here. If none, delete this.}
\thanks{\rev{This work has received funding from the Key Digital Technologies Joint Undertaking (KDT-JU) under grant agreement No 101095947. The JU receives support from the European Union’s Horizon Europe research and innovation programme.}}
}

% \author{\IEEEauthorblockN{1\textsuperscript{st} Given Name Surname}
% \IEEEauthorblockA{\textit{dept. name of organization (of Aff.)} \\
% \textit{name of organization (of Aff.)}\\
% City, Country \\
% email address or ORCID}
% \and
% \IEEEauthorblockN{2\textsuperscript{nd} Given Name Surname}
% \IEEEauthorblockA{\textit{dept. name of organization (of Aff.)} \\
% \textit{name of organization (of Aff.)}\\
% City, Country \\
% email address or ORCID}
% \and
% \IEEEauthorblockN{3\textsuperscript{rd} Given Name Surname}
% \IEEEauthorblockA{\textit{dept. name of organization (of Aff.)} \\
% \textit{name of organization (of Aff.)}\\
% City, Country \\
% email address or ORCID}
% \and
% \IEEEauthorblockN{4\textsuperscript{th} Given Name Surname}
% \IEEEauthorblockA{\textit{dept. name of organization (of Aff.)} \\
% \textit{name of organization (of Aff.)}\\
% City, Country \\
% email address or ORCID}
% \and
% \IEEEauthorblockN{5\textsuperscript{th} Given Name Surname}
% \IEEEauthorblockA{\textit{dept. name of organization (of Aff.)} \\
% \textit{name of organization (of Aff.)}\\
% City, Country \\
% email address or ORCID}
% \and
% \IEEEauthorblockN{6\textsuperscript{th} Given Name Surname}
% \IEEEauthorblockA{\textit{dept. name of organization (of Aff.)} \\
% \textit{name of organization (of Aff.)}\\
% City, Country \\
% email address or ORCID}
% }

\author{\IEEEauthorblockN{Daniele Jahier Pagliari\IEEEauthorrefmark{1}, Matteo Risso\IEEEauthorrefmark{1}, Beatrice Alessandra Motetti\IEEEauthorrefmark{1}, Alessio Burrello\IEEEauthorrefmark{1}}
\IEEEauthorblockA{\IEEEauthorrefmark{1}Politecnico di Torino, Turin, Italy.
}\textit{Corresponding Email: daniele.jahier@polito.it}
}
%\author{OMITTED FOR BLIND REVIEW}

\maketitle

\copyrightnotice

\begin{abstract}
\input{sec/00_abstract}
\end{abstract}


\input{sec/01_introduction}
\input{sec/02_background}
\input{sec/03_related_works}
\input{sec/04_method}
\input{sec/05_results}
\input{sec/06_conclusions}

%\bibliographystyle{IEEEtran}
%\bibliography{IEEEabrv,bib/bibliography}

\end{document}

%% file: sec/00_abstract.tex
Accurate yet efficient Deep Neural Networks (DNNs) are in high demand, especially for applications that require their execution on constrained edge devices.
Finding such DNNs in a reasonable time for new applications requires automated optimization pipelines since the huge space of hyper-parameter combinations is impossible to explore extensively by hand.
In this work, we propose PLiNIO, an open-source library implementing a comprehensive set of state-of-the-art DNN design automation techniques, all based on lightweight gradient-based optimization, under a unified and user-friendly interface.
\rev{With experiments on several edge-relevant tasks, we show that combining the various optimizations available in PLiNIO leads to rich sets of solutions that Pareto-dominate the considered baselines in terms of accuracy vs model size. Noteworthy, PLiNIO achieves up to 94.34\% memory reduction for a $<$1\% accuracy drop compared to a baseline architecture.}

%% file: sec/01_introduction.tex
\section{Introduction}
\rev{Deep Neural Networks (DNNs) reach state-of-the-art performance in many applications, ranging from computer vision to bio-signals processing, but are extremely expensive in terms of computation and memory~\cite{burrelloBioformers2022, liuBringing2022,Daghero2021a}}.
This is currently considered somewhat of a secondary issue for cloud-hosted models, whose accuracy has improved in each new generation as an effect of sheer model upscaling, also thanks to the availability of gargantuous amounts of data. However, for tasks that require the execution of DNNs on mobile or edge devices, limiting computational complexity and memory footprint is fundamental~\cite{liuBringing2022}, and even in the cloud, hardware/energy costs and sustainability issues will eventually mandate a careful consideration of complexity~\cite{an2023chatgpt}.

Unfortunately, DNNs have a very large set of hyper-parameters, i.e., configurations that are not (traditionally) trained by gradient descent together with the model weights, yet greatly influence results. At a high level, we can distinguish \textit{training hyper-parameters} (e.g., the optimizer used for training, the initial learning rate and its schedule, etc) and \textit{architectural hyper-parameters} (e.g., the number and type of layers, their configuration, the weights and activations bitwidth, etc). The former only affect the training process and, therefore, the accuracy of the resulting model. The latter, instead, strongly impact both predictive performance and inference complexity. Furthermore, they can be set in virtually infinite combinations, creating an immense optimization space~\cite{dongbench}. Exploring such space by hand is prone to following conventional rules of thumb and results in suboptimal outcomes~\cite{1000nas}.

Accordingly, \rev{design exploration and} automated optimization tools, generally referred to as AutoML~\cite{he2021automl} are becoming popular to design accurate yet compact and efficient DNNs for new applications, especially when targeting constrained edge hardware. More specifically, Neural Architecture Search (NAS) methods automate the search for optimal combinations of layers and their configurations~\cite{1000nas}, whereas Mixed-Precision Search (MPS) solutions look for the optimal data representation for each model tensor~\cite{wangHAQ2019}. In both cases, early approaches resorted to time-consuming black-box optimization methods such as Reinforcement Learning (RL) and Evolutionary Algorithms (EA), which
%
% iteratively trained and evaluated the accuracy and complexity of thousands of models, thus requiring
%
required
tens of GPU-days for a single optimization~\cite{1000nas}. More recently, gradient-based NAS and MPS have been proposed as lightweight alternatives to these solutions. These so-called \textit{One-shot} or \textit{Differentiable} methods utilize gradient-descent to simultaneously train a DNN and optimize its architecture, thus obtaining an optimized model in a time comparable to a single training~\cite{liu2018darts}.

One key limitation of gradient-based approaches, however, is the lack of user-friendly libraries that can be employed by ML practitioners without experience on NAS or MPS to optimize a DNN for their applications while ignoring implementation details. Such a library should also combine optimizations targeting multiple architectural hyper-parameters, at different granularity levels, in order to fully explore the design space. Similar tools have been recently released both commercially~\cite{edge_impulse} and open-source~\cite{autokeras}, but mostly for resource-hungry iterative (i.e., RL, EA, etc.) AutoML methods.

In this paper, we present \textbf{PLiNIO}, a library for \textbf{P}lug-and-play \textbf{Li}ghtweight \textbf{N}eural \textbf{I}nference \textbf{O}ptimization, which tries to bridge this gap by providing a \rev{unified and user-friendly domain-specific language for a diverse set of gradient-based AutoML procedures}. Namely, PLiNIO currently supports: i) a \textit{coarse-grained NAS} for selecting among alternative layers~\cite{liu2018darts}; ii) a \textit{fine-grained NAS} for optimizing each layer's internal hyper-parameters (e.g., the number of channels in a Convolutional layer)~\cite{pit_tcomp}; iii) a \textit{differentiable MPS} method for selecting both weights and activation bit-widths and quantization parameters, supporting common quantization formats~\cite{cai20edmips,pact,rissoChannelwise2022}. Since the fine-grained NAS in ii) is analogous to structured pruning~\cite{pit_tcomp}, PLiNIO supports three of the most common complexity-driven DNN optimizations in the state-of-the-art, i.e., \textbf{Quantization, Pruning and NAS}~\cite{liuBringing2022,Daghero2021a} under a unified API \rev{thus enabling a complete Design Space Exploration (DSE) for DNN workloads}. Furthermore, PLiNIO's internals are designed to support extensibility, and we plan to integrate the library with additional gradient-based optimizations in future releases. \rev{PLiNIO is available open-source at: \texttt{\url{https://github.com/eml-eda/plinio}}}.

\rev{Fig.~\ref{lst:api}} shows the modifications required to implement a PLiNIO optimization on top of a standard training loop in PyTorch, i.e., the DNN training framework on which our library is based. As shown, PLiNIO only requires three method invocations, highlighted in the figure, to convert a standard PyTorch DNN into an optimizable model \rev{(line 2)}, estimate its complexity \rev{according to one or more cost metrics} during the optimization \rev{(line 6)} and export the final optimized model at the end of the process \rev{(line 10)}. This interface applies to all supported gradient-based optimizations, making them very simple to integrate into existing codebases. 

To the best of our knowledge, PLiNIO is the first library to support gradient-based NAS, pruning and MPS in a single unified framework. In order to demonstrate its usefulness \rev{for DNN workloads' DSE}, we run experiments on three edge-relevant use-cases taken from the MLPerf Tiny benchmark suite~\cite{mlperf-tiny}, using corresponding state-of-the-art reference DNNs as starting point for the optimization. Our results show that by combining all three optimizations, PLiNIO can reduce the DNN storage size by up to 94.34\% with a limited accuracy drop (-0.92\%) with respect to the reference. Noteworthy, the final model has 78.88\% fewer parameters compared to the best one obtained applying \textit{each optimization individually}.
%
\begin{comment}
\begin{figure}[t]
  \centering
  \includegraphics[width=\columnwidth]{fig/train_loop_single.png}
  \caption{Standard PyTorch training loop turned into a PLiNIO optimization. The \texttt{Method()} call is a place-holder for \texttt{SuperNet()}, \texttt{PIT()} or \texttt{MPS()}.}
  \label{fig:api}
\end{figure}
\end{comment}
%
\begin{figure}[t]
\begin{mdframed}
\begin{std}{1}
model = MyNN()
\end{std}
\vspace{-0.45cm}
\begin{highlight}{2}
model = plinio.Method(model, {'cost': cost_fn}, ...)
\end{highlight}
\vspace{-0.45cm}
\begin{std}{3}
for epoch in range(N_EPOCHS):
  for sample, target in data:
    output = model(sample)
\end{std}
\vspace{-0.45cm}
\begin{highlight}{6}
    loss = criterion(output, target)+model.get_cost('cost')
\end{highlight}
\vspace{-0.45cm}
\begin{std}{7}
    optimizer.zero_grad()
    loss.backward()
    optimizer.step()
\end{std}
\vspace{-0.45cm}
% \begin{highlight_last}{11}{\rev{Standard PyTorch training loop turned into a PLiNIO optimization. The \texttt{Method()} call is a place-holder for \texttt{SuperNet()}, \texttt{PIT()} or \texttt{MPS()}.}}{lst:api}
\begin{highlight}{10}
exported_model = model.export()
\end{highlight}
\end{mdframed}
\vspace{-0.35cm}
\caption{{\rev{Standard PyTorch training loop turned into a PLiNIO optimization. The \texttt{Method()} call is a place-holder for \texttt{SuperNet()}, \texttt{PIT()} or \texttt{MPS()}.}}}\label{lst:api}
\vspace{-0.45cm}
\end{figure}

%% file: sec/02_background.tex
\section{Background}

\subsection{Neural Architecture Search}
%
% The burden of manually tuning the network’s hyperparameters in order to achieve the desired accuracy or hardware costs has highlighted a strong need to automate such design process. As a consequence, in recent years various Neural Architecture Search (NAS) tools have been proposed, which allow exploring a given architectural search space in order to find the best-performing network structure.
%
% NAS methods are built upon three main components~\cite{hw_nas_survey, 1000nas}: the \textit{search space}, the \textit{search algorithm}, and the \textit{evaluation strategy}. 
% %
% The \textit{search space} contains all the architectures and associated hyperparameters that we might want to explore, and can include as many as $10^{32}$ alternatives~\cite{pit_tcomp}.
%
% , which are represented as operations and connections between the latter. The bigger the search space, the more alternatives we can consider during the search, thus potentially reaching a greater performance at the end. However, such fine search granularity might be computationally unfeasible to handle, since the search space dimension can go beyond $10^{36}$~\cite{songhan_survey}. 
% o 10^32 di pit
%
% The \textit{search algorithm} defines the strategy for finding DNNs within this space that minimize a given objective function.

NAS search algorithms can be broadly categorized into black-box methods, such as RL and EA, and Differentiable NAS (DNAS) or one-shot methods. The former requires three main steps: i) sampling one or more architectures from the search space, ii) evaluating the objective function and then iii) updating the sampling policy. These methods can optimize almost any function, with both accuracy- and complexity-related terms, on a discrete search space~\cite{1000nas}. However, they are often intractable, mainly due to the evaluation step (ii).
In fact, evaluating the accuracy of a sampled DNN ideally requires training it to convergence, whereas cost metrics should be obtained directly from deployment, both of which largely increase the search time. This is partially mitigated by the use of ``proxies''~\cite{cai2018proxylessnas}, such as training on a subset of the dataset or for a few epochs~\cite{liu2018darts} and using Look-up Tables~\cite{cai2018proxylessnas} or other approximate models~\cite{liberis_nas_2021, snas} for cost metrics (e.g., memory occupation, latency or energy). Still, black-box methods remain extremely time-consuming~\cite{1000nas,songhan_survey}.

DNAS reduce the optimization time significantly by relaxing the search space from discrete to continuous, making the problem suitable for gradient-descent~\cite{1000nas}.
Namely, they define a set of \textit{architectural parameters} ($\theta$) which encode the selection of a DNN from the search space and train them together with the weights of the networks. 
When optimizing for both functional (accuracy) and non-functional metrics, a DNAS training loop typically uses a loss function in the form: 
\begin{equation} \label{eq:dnas}
\min_{W, \theta} \mathcal{L}(W; \theta) + \lambda \mathcal{R}(\theta)
\end{equation}
where $W$ are the normal DNN weights, $\mathcal{L}$ is the task-dependent functional loss, $\mathcal{R}$ is a differentiable cost estimate, and $\lambda$ is a strength hyper-parameter that controls the balance between the two. Typical expressions for $\mathcal{R}$ encode the model size (number of parameters) or inference operations (OPs) as a function of $\theta$, but more complex latency approximations are also possible~\cite{cai2018proxylessnas,multiloss_conf} \rev{(Sec.~\ref{sec:cost_models})}.
At the end of the search, the $\theta$ parameters are discretized to export the final DNN.

More specifically, \textit{path-based} DNAS methods~\cite{liu2018darts} define a DNN (the \textit{supernet}) whose graph includes multiple alternative paths corresponding to the possible alternative operations in the search space. 
The optimization reduces to selecting one of these paths, as detailed in Sec.~\ref{sec:path_based_dnas}.
%
% During training, the outputs of all paths are linearly combined, each weighted by an architectural hyperparameter. The training loop assigns larger values to path parameters that minimize the objective function. At the end, the operations associated with the largest architectural coefficients are selected. 
%
The main issue with this approach is that the supernet size grows quickly with the search space, limiting scalability. Advanced methods such as ProxylessNAS~\cite{cai2018proxylessnas} and HardCoReNAS~\cite{nayman_hardcore-nas_2021} solve this sampling few paths per training iteration.

\textit{Mask-based} DNAS further reduce the optimization cost by searching only among the DNNs that can be obtained by shrinking an initial architecture, called \textit{seed}, in a way similar to structured pruning. In particular, slices of the DNN weights or activations tensors are coupled with binary masks, whose continuous relaxation is trained with gradient descent. At the end of the search, the masked parts of the seed layers are eliminated. Thus, the search space of these methods is more restricted w.r.t. path-based DNAS (only subsets of the seed are explored). However, the search granularity can be much finer. Examples of \textit{mask-mased} methods are MorphNet~\cite{gordon2018morphnet}, FbNetV2~\cite{wan2020fbnetv2} and PIT~\cite{pit_tcomp}. 

\subsection{Quantization and Mixed-precision Search}

Integer quantization is a key DNN optimization, especially at the edge, consisting in the approximation of floating point weights and activations with low bitwidth integers, improving both model size and efficiency~\cite{jacob_quantization_2018}.
While the conversion can be done post-training~\cite{banner2019post_training_4bit_qtz }, simulating the effect of quantization at training time (so-called Quantization-Aware Training  or QAT)~\cite{jacob_quantization_2018} can help the DNN adapt to the data approximation, reducing the drop in accuracy.

Standard \textit{fixed-precision} quantization assigns the same integer bit-width to the whole DNN, thus neglecting the sensitivity of each layer to precision reductions. However, previous works~\cite{cai20edmips, rissoChannelwise2022} show that some layers (e.g., those close to the input and to the output) tend to require higher precision. \textit{Mixed-precision} methods address this issue by quantizing various subsets of the DNN at different bitwidths. This creates a new and not trivial optimization problem, i.e., finding precision assignments that yield good trade-offs between accuracy and complexity, exploring a search space whose size increases exponentially with the number of considered bitwidths~\cite{dong_hawq_2019}.

Various MPS approaches have been proposed to tackle this problem, which is orthogonal to NAS. Some exploit sensitivity metrics such as the layers' Hessian spectrum~\cite{dong_hawq_2019} or the Signal to Quantization Noise Ratio at different precisions\cite{pandey_practical_2023}, while others are based on RL~\cite{wangHAQ2019}. More recently, the authors of~\cite{cai20edmips,rissoChannelwise2022} proposed a gradient-based method similar to DNAS to assign bitwidths during training. This is done by quantizing data at every possible precision on-the-fly, and then learning to select a single precision during training, similarly to~\cite{liu2018darts}.

%% file: sec/03_related_works.tex
\section{Related Works}
\label{sec:related}
\begin{comment}
%
\textbf{@MATTE?}
%
\textbf{This should focus on existing NN Optimization libraries. This is one the most tricky parts to write, needs an updated review of what is out there. Mentioning:
\begin{itemize}
    \item TFLite and TFMOT
    \item PyTorch Quantization (and other opt. if any)
    \item Microsoft NNI
    \item AutoML tools like Edge Impulse and Qeexo
    \item Others?
\end{itemize}}
\end{comment}
%
The need for efficient and accurate DNNs has brought a plethora of techniques and algorithms for AutoML, including  NAS, pruning and quantization~\cite{1000nas}.
Given the very quick innovation pace, the software ecosystem is naturally fragmented, with most techniques being shared as isolated and hardly usable ``research scripts''.
Recently, comprehensive and engineered tools have started to emerge, both commercially and open-source, to tackle this problem.

Commercial tools for DNN optimization are usually inserted in larger AutoML pipelines, which also help users with data preprocessing and labeling, model deployment, etc, and are commonly offered as cloud services.
Some examples include
Azure Machine Learning%~\cite{azureML}
, Google Cloud AutoML%~\cite{google_automl}
, etc.
Other providers offer similar features but target specifically constrained edge devices.
Qeexo~\cite{qeexo_automl} is an example of no-code end-to-end autoML platform for the edge, in which model selection is performed by selecting a specific instance from a zoo of predefined algorithms, which are then quantized at fixed precision.
Edge Impulse's EONTuner~\cite{edge_impulse} follows a similar approach but offers more flexibility for model selection, with the possibility to define a coarse search space (e.g., number of convolutional/linear layers, presence of pooling, etc), searched with hyperband or random-search.

In the open-source landscape, DL frameworks such as Tensorflow and PyTorch support basic optimizations natively.
Namely, the TensorFlow Model Optimization Toolkit (TFMOT)~\cite{tfmot} implements fixed-precision QAT, pruning, and weight clustering, generating models compatible with the TFLite converter and interpreter for edge devices.
Likewise, PyTorch supports different quantization and pruning techniques and it exposes APIs to implement new ones~\cite{pytorch}.
Similar optimizations are also targeted by the open-source AI Model Efficiency Toolkit (AIMET)~\cite{aimet}.

Concerning NAS and hyperparameters optimization, one of the first attempts to realize a user-friendly library is represented by AutoKeras~\cite{autokeras}. This tool lets users define search spaces or provides pre-defined ones for specific tasks such as Image or Text Classification. It then offers different search strategies, including hyperband and Bayesian optimization. 
Neural Network Intelligence (NNI)~\cite{nni2021} by Microsoft implements several optimization algorithms under a unified framework with a common API. In particular, it supports Bayesian and heuristic-based hyper-parameters optimization, NAS with both iterative and gradient-based approaches, pruning, and fixed-precision quantization.
Vega~\cite{vega} similarly groups in the same codebase different NAS algorithms along with pruning and mixed-precision assignment based on EA.
To the best of our knowledge, no existing single framework supports path-based DNAS, mask-based DNAS, and gradient-based MPS with a unified interface.

%% file: sec/04_method.tex
\section{PLiNIO}
PLiNIO is, to our knowledge, the first open-source tool to support multiple gradient-based DNN optimizations, spanning the dimensions of: i) coarse architectural choices (path-based DNAS), ii) layer hyper-parameters optimization (mask-based DNAS) and iii) precision selection (MPS), through a simple, user-friendly interface. Its main scientific value is allowing to study the combination of \rev{DSE and} optimizations at different levels, and their interactions, which may lead to superior results with respect to any single method (see Section~\ref{sec:results}).

While there exist tools with similar flexibility leveraging at least in part black-box methods~\cite{vega}, an entirely gradient-based toolchain helps to democratize research in this field, since lightweight gradient-based methods \rev{might be the \textit{only alternative} to being forced to use third-party cloud services for users that do not own large GPU clusters}.
One domain where this is particularly relevant is TinyML~\cite{Daghero2021a}, i.e.,  systems that implement DNN inference directly on tightly constrained mobile or IoT edge devices. Although PLiNIO can in principle support the optimization of any DNN, regardless of its size and of the hardware target, TinyML is in fact its primary use case.

In the rest of this Section, we first describe the three optimization techniques currently supported by the library (Sec.~\ref{sec:supported_methods}). We then detail some of the DNN graph transformation passes required to implement the simple interface of \rev{Fig.~\ref{lst:api}}, hiding most complexity from the users (Sec.~\ref{sec:conversion}). Lastly, we discuss the extensible DNN cost models supported by PLiNIO for complexity-aware optimization (Sec.~\ref{sec:cost_models}).

\subsection{Supported Optimization Techniques}\label{sec:supported_methods}
\subsubsection{SuperNet}\label{sec:path_based_dnas}
\begin{figure}[t]
  \centering
  \includegraphics[width=0.9\columnwidth]{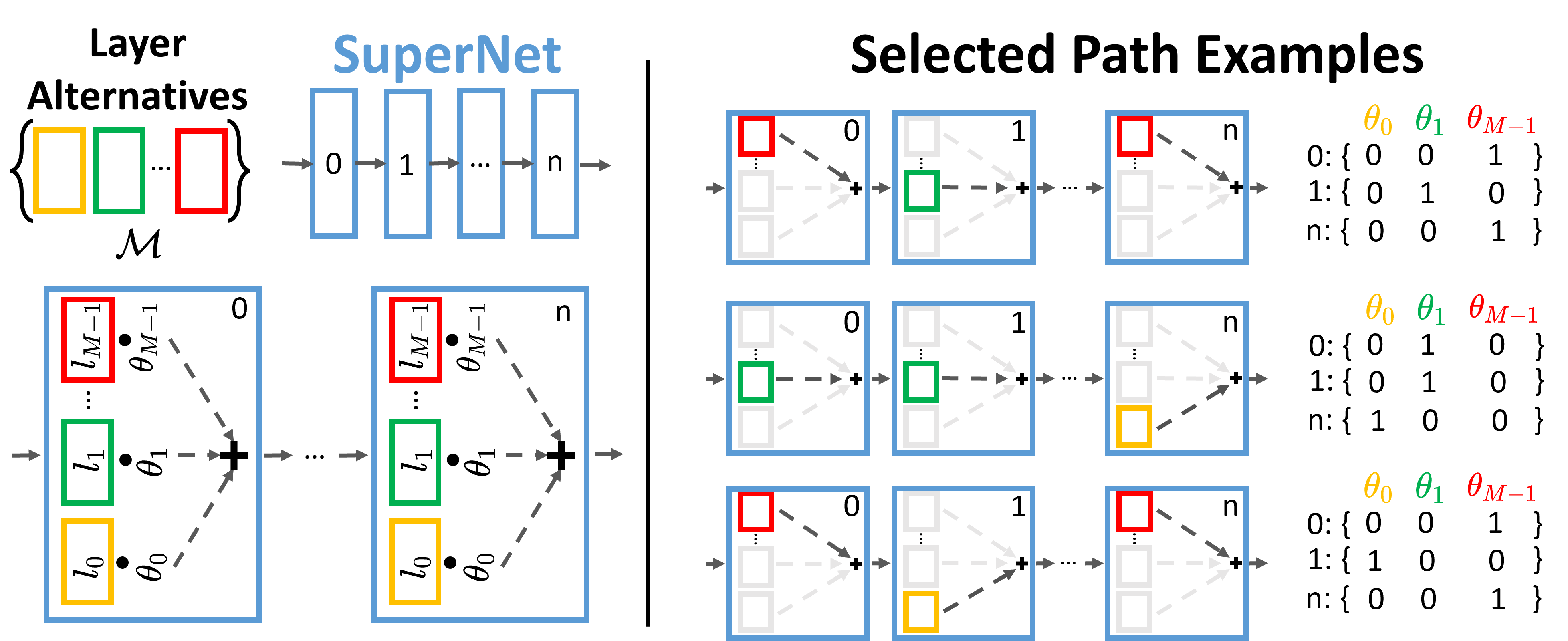}
  \vspace{-0.25cm}
  \caption{SuperNet implementation in PLiNIO.}
  \label{fig:path_dnas}
  \vspace{-0.4cm}
\end{figure}
%
% \textbf{Describe the DARTS-like method that we use, with differences w.r.t. the original one (e.g. Gumbel).}
%
As the first DNAS, PLiNIO implements a \textit{path-based} method based on a \textit{supernet}, schematized in Fig.~\ref{fig:path_dnas}.
The method is inspired by DARTS~\cite{liu2018darts}, but it differs by using the Gumbel-Softmax sampling strategy, in accordance with more recent works~\cite{snas, wan2020fbnetv2}, instead of the standard SoftMax as done in~\cite{liu2018darts}.
The supernet is built by replacing each layer $L$ of a standard DNN with an ensemble of $M$ alternatives $\mathcal{M} = \{l_{i}\}_{m=0}^{M-1}$.
All $l_{i} \!\in\! \mathcal{M}$ receive the same input, and their outputs are linearly combined using trainable parameters $\theta_{i}$, passed through a Gumbel-Softmax ($GS$), i.e.,:
\begin{equation}\label{eq:supernet}
Y = \sum_i GS(\theta_{i}) \cdot l_i(X)
\end{equation}
The selection of layers is reduced to training $\theta$, jointly with the normal layer weights $W$, to minimize Eq.~\ref{eq:dnas}.
At the end of the training, the optimized DNN is obtained by selecting, from each set, the alternative corresponding to the largest $\theta_i$.
\subsubsection{PIT}
\begin{figure}[t]
  \centering
  \includegraphics[width=0.9\columnwidth]{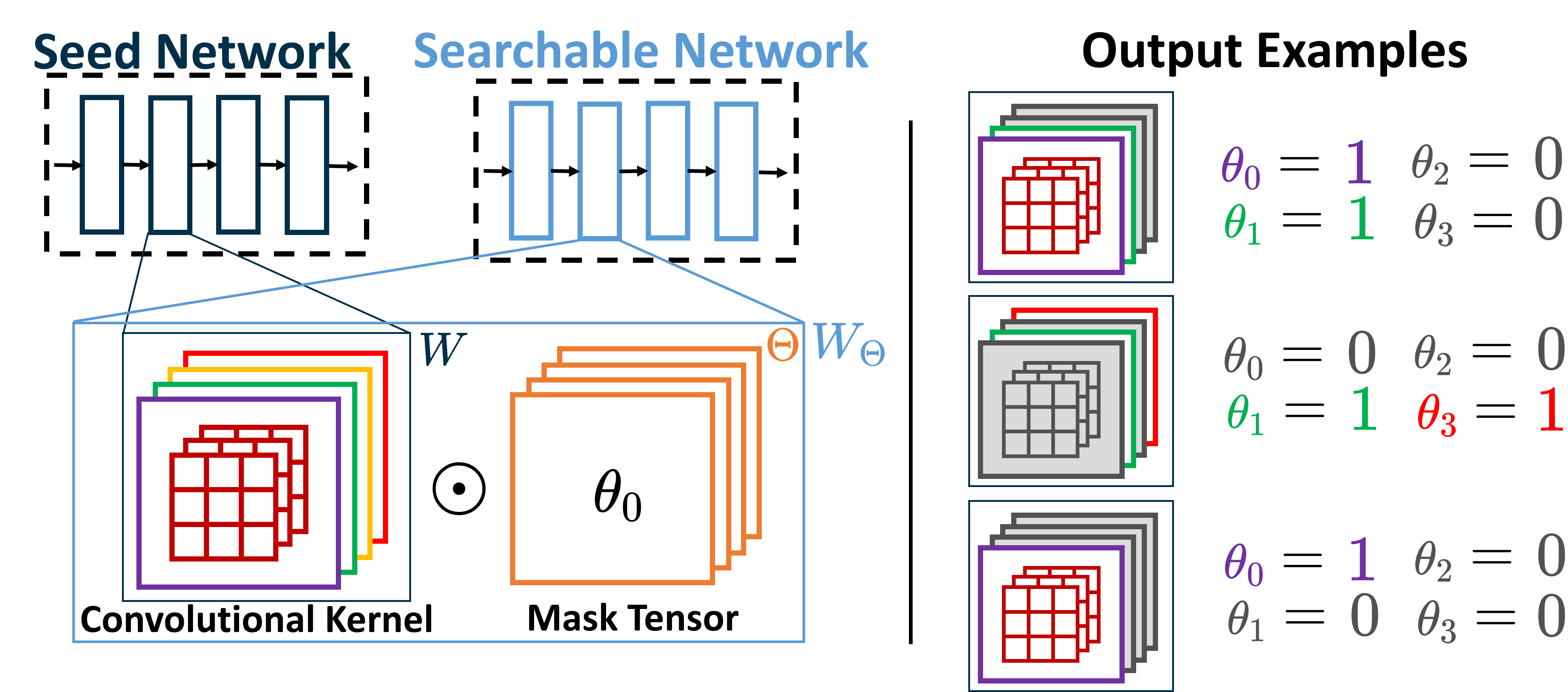}
  \vspace{-0.25cm}
  \caption{PIT channel-masking implementation in PLiNIO.}
  \vspace{-0.25cm}
  \label{fig:mask_dnas}
\end{figure}
%
% \textbf{Describe briefly PIT, focusing on the channel pruning part, and mentioning that it can also affect RF and Dilation, referring to the TCOMP journal for details. }
%
The second DNAS implemented in PLiNIO is a \textit{mask-based approach} akin to structured pruning. Starting from a seed network, this method optimizes the main architectural hyper-parameters of Convolutional (Conv) and Fully-Connected (FC) layers at fine grain.
It extends PIT~\cite{pit_tcomp}, which was originally proposed to optimize the most important hyper-parameters of 1D Conv, (i.e., number of output channels, receptive-field, and dilation) to also support output channels optimization for 2D layers.

Its channel-search approach is summarized in Fig.~\ref{fig:mask_dnas}.
Starting from the seed, each Conv or FC weight tensor $W$, with $C_{out}$ output channels, is masked as follows:
\begin{equation} \label{eq:mask_ch}
W_{\Theta} = W \odot \mathcal{H}(\theta)
\end{equation}
where $\theta$ is a vector of $C_{out}$ trainable mask parameters, $\odot$ is the Hadamard product, and $\mathcal{H}$ is a Heaviside step function to binarize $\theta$.
Each element $\theta_{i}$ masks an entire output channel of $W$, controlling whether it is kept ($\mathcal{H}(\theta_i) = 1$) or removed from the network ($\mathcal{H}(\theta_i) = 0$).

Similarly to the supernet of Sec.~\ref{sec:path_based_dnas}, \rev{the DNN with masked weights is inserted in a normal training loop}, where $W$ and $\theta$ are trained together to minimize Eq.~\ref{eq:dnas}.
During forward training passes, the use of $\mathcal{H}$ has the effect of sampling of one architecture from the search space, as shown on the right of Fig.~\ref{fig:mask_dnas}.
Instead, in backward passes, a Straight-Through Estimator (STE)~\cite{pit_tcomp} technique is used to ensure that gradients flow through the non-differentiable $\mathcal{H}$.
\subsubsection{MPS}
\begin{figure}[t]
  \vspace{-0.25cm}
  \centering
  \includegraphics[width=0.9\columnwidth]{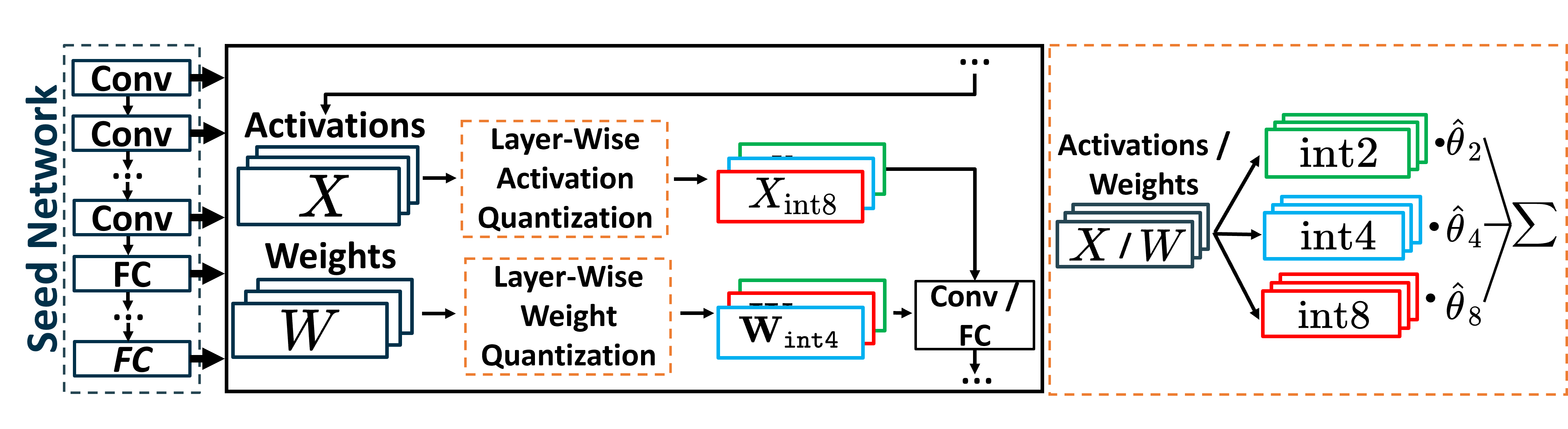}
  \vspace{-0.25cm}
  \caption{MPS implementation in PLiNIO.}
  \label{fig:mixprec_dnas}
  \vspace{-0.4cm}
\end{figure}
Fig.~\ref{fig:mixprec_dnas} summarizes the MPS method implemented in PLiNIO to assign independent precision to weights $W$ and activations $X$ in  Conv and FC layers.
The method is inspired by~\cite{cai20edmips}, extended with additional quantization formats. 
Given the set of supported bit-widths $p \in P$, e.g., $P = \{2, 4, 8\}$, each tensor $T$ (either $W$ or $X$) is fake-quantized~\cite{jacob_quantization_2018} at all bit-widths. The differently quantized ``variants'' are
linearly combined by trainable parameter vectors $\theta$ of length $\|P\|$, normalized by means of a SoftMax function ($SM$).
In practice, an \textit{effective} tensor is obtained as:
\begin{equation}\label{eq:mps_layer}
\hat{T} = \sum_{p} SM(\theta_{p}) \cdot T_{p}
\end{equation}
where $T_{p}$ is the $p$-bit version of $T$.
Therefore, increasing the value of $\theta_{p}$ causes the output tensor $\hat{T}$ to resemble more the result of $p$-bit quantization.
Importantly, all fake-quantized versions are derived from a \textit{single} float tensor, thus minimizing the method's memory overhead at training time.

The effective tensors $\hat{W}$ and $\hat{X}$ are then used to compute the layer's output, e.g.,: $Y = \mathrm{Conv}(\hat{X}, \hat{W})$.
As for the other PLiNIO methods, the DNN, thus modified, is inserted in a DNAS-like training loop to jointly optimize $W$ and $\theta$ according to Eq.~\ref{eq:dnas}.
\subsection{Automatic Model Conversion and Export}\label{sec:conversion}

\begin{figure}[t]
  \centering
  \includegraphics[width=.95\columnwidth]{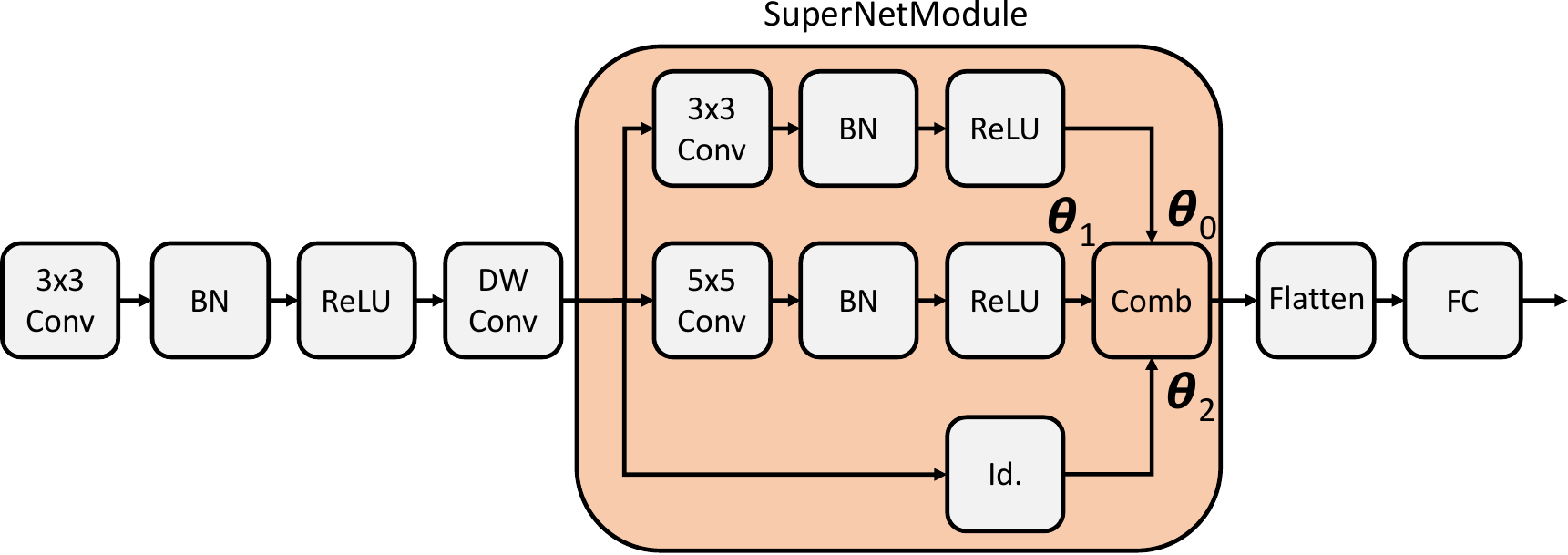}
  \vspace{-0.3cm}
  \caption{SuperNet module example. Id. = Identity operation.}
  \label{fig:sn}
  \vspace{-0.4cm}
\end{figure}

PLiNIO lets users define the optimization input DNN as a standard \texttt{nn.Module} sub-class, as in vanilla PyTorch.
The only special DNN definition construct is a new type of ``layer'', through which users can explicitly define the alternative paths that form the search space of the method in Sec.~\ref{sec:path_based_dnas}. The constructor of this class, called \texttt{SuperNetModule}, takes as input a list of \texttt{nn.Module} instances, each corresponding to a possible optimization alternative, i.e., either a single layer or a more complex sub-network. Fig.~\ref{fig:sn} shows an example with three inputs, where the additional Comb node, \rev{added automatically during the conversion}, combines the various branches through Eq.~\ref{eq:supernet}. This  gives maximum freedom to users, allowing them to easily consider different alternatives for each layer rather than a fixed set of operations for the whole DNN.
Besides this, all other transformations required to make a standard PyTorch DNN optimizable by PLiNIO occur \textit{transparently}, when the model is passed to a method's constructor \rev{(line 2 of Fig.~\ref{lst:api})}. Namely, a series of conversion passes are performed, which make extensive use of the \texttt{torch.fx} toolkit, as detailed below. Fig.~\ref{fig:conversion} shows an example of this conversion for PIT on a portion of a plausible DNN graph.
\subsubsection{Layer Auto-conversion}\label{sec:layer_conv}
PIT and MPS are commonly applied to \textit{all} Conv and FC layers of a DNN. Thus, in this case, PLiNIO does not require users to explicitly define optimizable layers. Rather, it identifies and converts \texttt{nn.Conv} and \texttt{nn.Linear} layers automatically (orange boxes in Fig.~\ref{fig:conversion}), adding architectural masks/parameters ($\theta$) as needed. Optional user-specified rules, by name or type, can exclude parts of the model from the optimization.

One key analysis pass performed during layer conversion is \textit{mask sharing}. In fact, for both PIT and MPS, the architectural parameters of different layers shall not always be optimized independently of each other. For instance, each output channel of the DepthWise (DW) Conv layer in Fig.~\ref{fig:conversion}b processes a different input channel. Therefore, if the $j$-th output channel is masked by PIT based on Eq.~\ref{eq:mask_ch}, the $j$-th input activation map also becomes useless, and so do the weights and computations of the preceding 3x3 Conv layer \rev{that produced it.}
%
% are practically wasted.
This is addressed \textit{sharing the channels mask} for the two layers ($\theta_A$ in the figure). Similar reasoning also applies to the two 1x1 Conv layers, whose outputs converge into an element-wise Add.
%
% Since the latter requires identically-shaped inputs, channels in corresponding positions should always be pruned in both Conv layers.
%
MPS auto-conversion includes an analogous pass for layers that need to share the same output activations bit-width and quantizer parameters, such as the two Add inputs. Tools such as EdMIPS~\cite{cai20edmips} and the NAS API of NNI~\cite{nni2021} do not apply mask sharing automatically, producing outputs that require further post-hoc transformations to become deployable, possibly affecting both their accuracy and their cost.

\subsubsection{Batch Normalization Folding}
\begin{figure}[t]
  \centering
  \includegraphics[width=1\columnwidth]{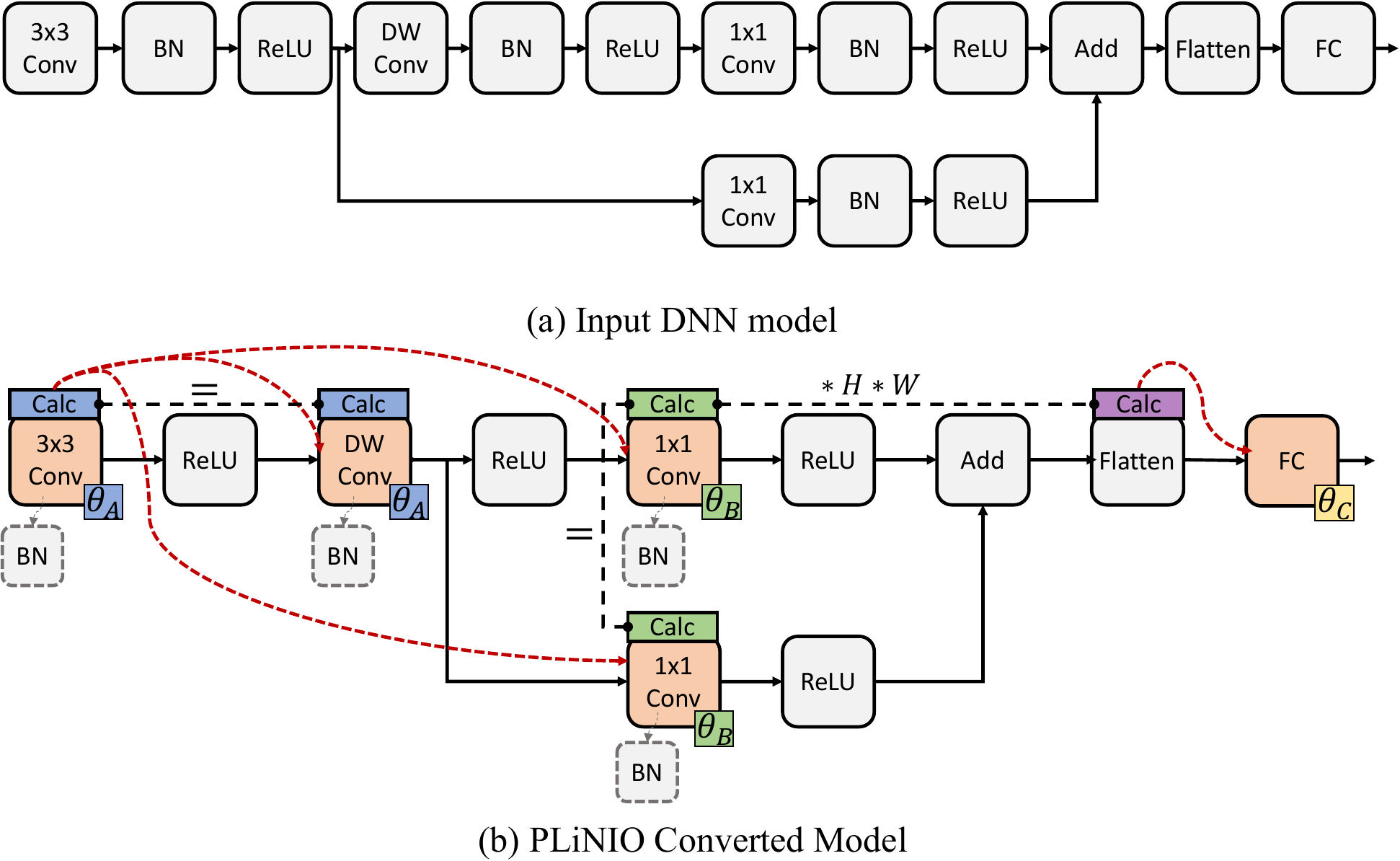}
  \vspace{-0.4cm}
  \caption{PLiNIO auto-conversion example.}
  \vspace{-0.45cm}
  \label{fig:conversion}
\end{figure}

For PIT and MPS, another pass folds Batch Normalization (BN) with the preceding Conv/FC layers. For MPS, this is needed to closely mimic a full-integer inference, which normally does not support BN, thus improving the consistency between the fake-quantized layers and the final integer model.
For PIT, folding is required because the zero-mean output of BN would back-propagate a very small-magnitude gradient from the task loss term $\mathcal{L}$ of Eq.~\ref{eq:dnas} to the channel mask parameters of Eq.~\ref{eq:mask_ch}. Thus, the masks gradients would be dominated by the cost term $\mathcal{R}$. In other words, the optimization would prune channels only based on their cost and not on their impact on accuracy.
PLiNIO saves the original BN parameters in a special Conv/FC field (small dashed squares in Fig.~\ref{fig:conversion}b), permitting to optionally \textit{unfold} the BN at the end of the optimization.
%
% Indeed, since the output of a PIT optimization is still a floating point model, users might want to restore BN, e.g., before performing fine-tuning.

%
\subsubsection{Effective Input Shape/Bitwidth Calculation}\label{sec:input_shape}
A layer's input tensor shape and precision greatly influence its cost. For instance, pruning some channels from the 1x1 Conv layers in Fig.~\ref{fig:conversion}b with PIT not only reduces their size/OPs, but also affects the cost of the following FC layer, which has to process a smaller number of inputs.
As for architectural parameters sharing, many other libraries do not account for these cost dependencies during gradient-based optimization.

\looseness=-1
PLiNIO does so by first performing a DNN graph traversal that associates each layer to the one(s) that determine its input tensor shape/bitwidth. An example of the result for the \textit{channels} dimension is shown by red dashed lines in Fig.~\ref{fig:conversion}b. The map is created for all nodes but shown only for the orange ones for simplicity. In practice, the \rev{static analysis} pass identifies layers that may alter the number of channels (e.g., 3x3 or 1x1 Conv) or not (e.g., DW Conv or ReLU), \rev{hierarchically traversing user-defined \texttt{nn.Module}s}. It then associates each layer with its closest channel-defining predecessor. Special cases are also dealt with, e.g., concat operations. %, whose number of output features depends on all their inputs.
A similar association is also done for MPS, linking each layer with the one that determines its input activations bitwidth. %(e.g. a ReLU is not \textit{bitwidth-defining} since it does not alter the number of bits of its input). 

Then, a second pass associates \textit{effective shape/bitwidth calculator} objects to each layer. The effective shape differs from the static size of PyTorch tensors because, for example, PIT does not actually \textit{eliminate} parts of the layer but only sets them to zero. 
Thus, the static shape remains unchanged until the final DNN export,
and using it to estimate a layer's computational cost, e.g., in terms of parameters or OPs would lead to gross over-estimations.
Effective shape calculators (Calc), shown as small coloured rectangles in Fig.~\ref{fig:conversion}b for the channels dimension, solve this issue by estimating the shape that would be obtained by exporting the currently sampled model as a function of the $\theta$ parameters. For example, if a binarized $\theta$ array for a layer with $C_{out} = 32$ has 20 zeroes, then $C_{out,eff} = 12$. \rev{Clearly, layers that share the same masks also share the calculator.} Additionally, more complex relationships are also inferred. For instance, the number of output channels of the Flatten operation in Fig.~\ref{fig:conversion}b depends on the preceding Conv through a multiplicative factor.
A similar mechanism estimates the \textit{effective input bitwidth} for MPS, since the input activations precision is relevant for estimating the time/energy cost of a layer.
\subsubsection{Final Model Export}
At the end of a PLiNIO search loop, the \texttt{export()} method \rev{(line 10 of Fig.~\ref{lst:api})} triggers an opposite conversion process to output the final optimized model as a vanilla \texttt{nn.Module}, that users can further train or deploy using their existing infrastructure.

To this end, the model is cleaned up from all the support structures added by PLiNIO, and the target layers are converted back to the corresponding standard PyTorch classes. \texttt{SuperNetModule} instances are replaced by the selected branches, and the combiner node is removed from the DNN graph. PIT target layers are replaced with a new instance of the same type which does not include the pruned portions, and the weights which have been preserved by the optimization are copied to the new layer. BN is optionally unfolded. For MPS, all layers are converted to fake-quantized versions at the selected precision. This serves as an intermediate step that allows us to possibly \rev{fine-tune} the model before the final integerization, which is hardware-specific~\cite{jacob_quantization_2018}.

\subsection{DNN Cost Specification}\label{sec:cost_models}
\begin{figure}[t]
  \centering
  \includegraphics[width=\columnwidth]{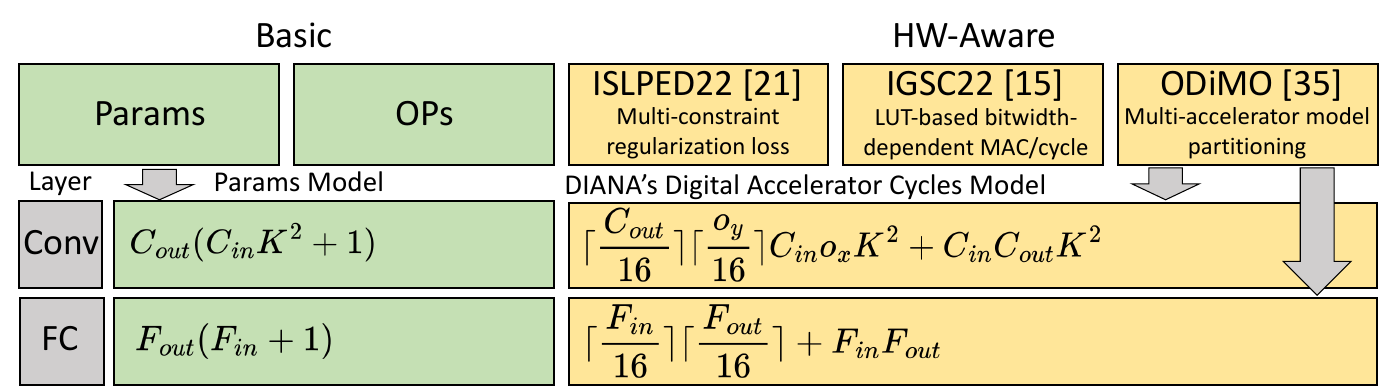}
  \caption{\rev{PLiNIO cost specification examples.}}
  \vspace{-0.25cm}
  \label{fig:cost_fn}
  \vspace{-0.4cm}
\end{figure}

PLiNIO is flexible with respect to the definition of the DNN complexity model used for the optimization ($\mathcal{R}$ in Eq.~\ref{eq:dnas}). Default cost metrics such as model size or number of OPs are provided with the library, \rev{as well as more detailed models for specific HW targets. Some examples, reported in Fig.~\ref{fig:cost_fn}, include the LUT-based bitwidth-dependent MAC/cycle model for the MPIC RISC-V processor's vector unit of~\cite{rissoChannelwise2022} and the dataflow-aware cycles model for the accelerators of~\cite{odimo}.
A dictionary of cost specifications is passed to the PLiNIO constructors (line 2 of Fig.~\ref{lst:api}), and the corresponding cost values as a function of the optimization parameters $\theta$ can then be retrieved using the \texttt{get\_cost} method of the optimizable model (line 6).} We support more than one cost specification, \rev{which the user can freely combine through any differentiable function when computing the total loss,} to allow DNN optimization over multiple complexity dimensions, for instance, trying to balance accuracy and latency under a size constraint, as discussed in~\cite{multiloss_conf}.

Cost specifications are key-value maps, associating DNN graph patterns to differentiable PyTorch functions that estimate the corresponding cost as a scalar. A simple example is shown in Fig.~\ref{fig:cost_fn} for the \rev{``params'' model, and for the digital accelerator model of~\cite{odimo}}. The parameters passed to the cost function include all geometrical shapes of the matched layers for PIT and SuperNet (e.g., for a Conv., input/output channels, kernel size, dilation, etc.), and the bit-width of all involved tensors (inputs, weights, biases and outputs) for MPS.

User-defined cost specifications can use all or a subset of these inputs, depending on their level of detail. Inputs use the default naming of PyTorch (e.g., $C_{out}$ is \texttt{out\_channels} for a Conv), to make the definition of cost metrics as orthogonal as possible to the optimization method. It is then PLiNIO's responsibility to internally invoke the function with the correct input values. For example, PIT will substitute the static $C_{out}$ with \rev{$C_{out,eff}(\theta)$} calculated as discussed in Sec.~\ref{sec:input_shape}.
Cost specifications also provide a default behaviour for unmatched DNN graph portions, which is often to assume 0 cost (e.g., for negligible operations) or to trigger an exception.

%% file: sec/05_results.tex
\begin{figure*}[t]
  \centering
  \includegraphics[width=0.9\textwidth]{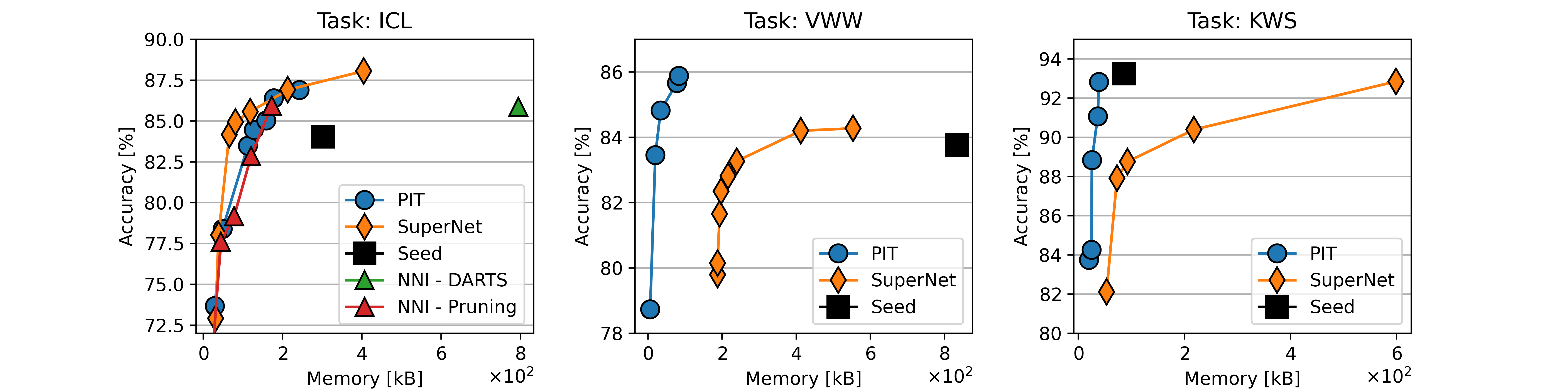}
  \vspace{-0.25cm}
  \caption{\rev{Application of PIT and SuperNet algorithms from the PliNIO library to three benchmarks of the MLPerf Tiny Suite.}}
  \label{fig:results_pit_supernet}
  \vspace{-0.4cm}
\end{figure*}
\input{sec/10_table1}
\section{Results}\label{sec:results}
\subsection{Experimental Setup}
We test PLiNIO on three benchmarks taken from the MLPerf Tiny suite~\cite{mlperf-tiny}.
Namely, image classification (ICL), visual wake word (VWW), and keyword spotting (KWS).
For each task, the suite defines a reference DNN, which we use as seed for PIT or as blueprint to construct the SuperNet.
We create the SuperNet replacing all Conv layers of the reference DNN with a \texttt{SuperNetModule} that selects between: i) a Conv with $3 \times 3$ filter, ii) a Conv with $5 \times 5$ filter, iii) a DW-separable convolution, which consists of a $3 \times 3$ DW Conv followed by a $1 \times 1$ pointwise Conv~\cite{mlperf-tiny} and iv) an identity operation, to possibly skip the layer.

The ICL task is based on the CIFAR-10 dataset and the reference DNN is a ResNet-like architecture with 8 Conv layers.
For VWW, the goal is to classify whether an input image contains at least one person. The dataset is MSCOCO 2014 with a reference model based on MobileNetV1 with a width multiplier of 0.25.
Lastly, KWS uses the Speech Commands v2 dataset.
The reference architecture is a simple DW Separable CNN (DS-CNN)\cite{mlperf-tiny}.
\rev{For sake of space, we omitted the least interesting MLPerf Tiny task, Anomaly Detection, whose reference architecture is a fully-connected autoencoder which does not offer the possibility to explore different layer alternatives with SuperNet.}
PLiNIO is implemented using Python 3.9 and PyTorch 1.13.1.
We compare the results of PLiNIO optimizations with the reference DNNs for each task.
{\color{blue}
%We cannot compare with other gradient-based AutoML tools, since as discussed in  Sec.~\ref{sec:related}, none of them supports the entire set of optimizations. Comparing with NNI\cite{nni2021}, for instance, would be unfair since it does not support MPS.
}
\rev{Moreover, we compared the SuperNet and PIT results on ICL with two similar methods taken from NNI~\cite{nni2021} 2.10.1.}
\rev{All experiments are executed on a machine with 32 GBs of RAM, a Intel(R) Core(TM) i3-9100F CPU running at 3.60GHz, and a Quadro P2200 GPU.}
% Further, their only supported mask-based DNAS is FBNetV2~\cite{wan2020fbnetv2}, which has been shown to be inferior to PIT in~\cite{pit_tcomp}.
%
We use the model size as PLiNIO's cost model for \rev{complexity}-aware optimization (see Sec.~\ref{sec:cost_models}).
%
%Similar results with other cost models (e.g. for latency or energy) are possible but not reported due to space limitations.

\subsection{Single NAS: PIT vs SuperNet}
\label{sec:res_global}
\rev{The results of applying PIT and SuperNet individually to the three reference DNNs are shown in Fig.~\ref{fig:results_pit_supernet}.} Each plot shows the reference (black square) and the optimized architectures (coloured dots) in the accuracy vs model size. The $\lambda$ in Eq. \ref{eq:dnas} was varied in the range between 1e-2 and 1e-10. 

The left plot shows that, on ICL, SuperNet tends to outperform PIT in terms of accuracy for a given storage footprint budget. Table \ref{tab:1} reports some of the most interesting architectures found by PLiNIO, whose memory spans from 31.6 kB to 405.08 kB and accuracy ranges from 72.9\% to 88.05\%. SuperNet achieves the highest accuracy (4.02\% higher than the seed) with a memory overhead of 34.04\%. Additionally, SuperNet also achieves the greatest reduction in memory (73.23\%) without accuracy loss (+0.91\%). 

The middle graph shows that, on the contrary, PIT greatly outperforms SuperNet on VWW. This is due to the large number of channels in seed layers, which creates a lot of memory-saving opportunities for mask-based DNAS, and demonstrates the importance of having both types of optimization in the library.
Many of the discovered architectures Pareto-dominate the seed: PIT finds DNNs that achieve between 78.73\% and 85.88\% accuracy with memory between 6.24 kB and 83.55 kB. \rev{At Iso-Accuracy, we achieve a striking 97.60\% memory reduction.}
The SuperNet approach, while being outperformed by PIT, is still capable of extracting architectures that are smaller yet equally accurate than the seed. 

The right plot shows the results on KWS, where PLiNIO finds many Pareto-optimal solutions with the PIT algorithm, spanning between 83.74\% and 92.82\% accuracy. Conversely, SuperNet never outperforms the seed, a result that testifies the goodness of the hand-tuned layer selection for this particular reference DNN. Similar to VWW, the networks found with SuperNet are outperformed by those found with PIT, indicating that changing the layer types or removing some of them is not beneficial for all tasks.

\rev{The time to complete one search epoch with SuperNet and PIT is comparable to one training epoch of the reference DNN.
For instance, on ICL, one PIT epoch is 1.8$\times$ slower than the reference, whereas SuperNet takes 5$\times$ longer, due to replacing each layer with multiple alternatives.}

\rev{The left-most graph of Fig.~\ref{fig:results_pit_supernet} also compares SuperNet and PIT with two similar approaches from NNI~\cite{nni2021}. The comparisons are presented only on ICL for the sake of space.}
\rev{In particular, SuperNet is compared with NNI's \textit{GumbelDARTS} method (green triangle). Despite exploring the same search-space, GumbelDARTS obtains a single optimized DNN, since it does not support complexity-driven search. The obtained point is strongly outperformed by PLiNIO's SuperNet, with $4\times$ size reduction at Iso-Accuracy. From a training-time perspective, PLiNIO is also 11\% faster than NNI.}
\rev{PIT, instead, is compared with NNI's \textit{L1NormPruner}, which again supports the same search-space. The two obtained Pareto-curves are very similar, with PIT slightly outperforming NNI by up to +0.64\% accuracy at Iso-Memory. Further, one key drawback of the NNI pruning is that it poorly supports complexity awareness, only allowing users to set the sparsity of individual layers (a knob difficult to control to achieve, for instance, a target model size) while not supporting other cost models.}

\subsection{Combination of NAS: SuperNet $\rightarrow$ PIT}
\begin{figure}[t]
  \centering
  \includegraphics[width=0.99\columnwidth]{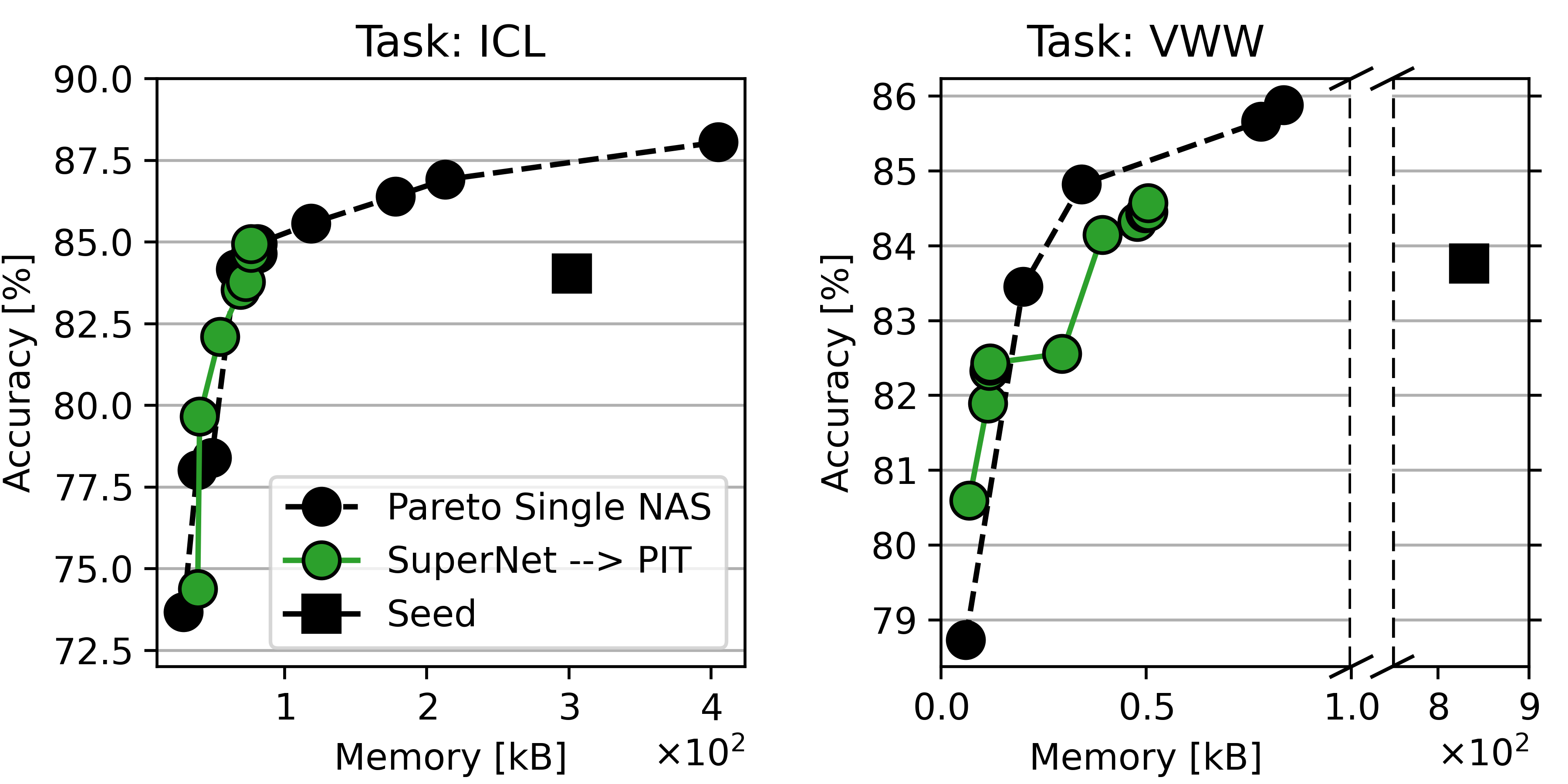}
  \vspace{-0.4cm}
  \caption{\rev{Comparison between the application of a single NAS and the concatenation of SuperNet and PIT.}}
  \label{fig:results_pit+supernet}
  \vspace{-.4cm}
\end{figure}

Fig.~\ref{fig:results_pit+supernet} depicts the results obtained by sequentially applying the two NAS algorithms in PLiNIO. The rationale is to first select the optimal number and type of layers with SuperNet, then optimize each layer's hyper-parameters at fine-grain with PIT.
\rev{In this case, the total cost is the sum of the cost of the two optimizations.}
\rev{We test this on the two benchmarks for which the Supernet approach had identified networks that either outperformed the seed in accuracy, or achieved Iso-Accuracy with a smaller model size, i.e., ICL and VWW.}
\rev{We obtain the green curves exporting the solutions found by SuperNet (line 10 of Fig.~\ref{lst:api}) to standard PyTorch networks and, then, applying PIT to the SuperNet results reported in Table~\ref{tab:1}.}
\rev{The black baseline curve is the combination of the two Pareto fronts of Fig.~\ref{fig:results_pit_supernet} obtained by applying PIT (blue curve) and SuperNet (orange curve) independently.} 

On both benchmarks, applying the two techniques in sequence results in small yet not negligible memory reductions. For instance, on ICL, we further reduce the memory usage by 4.5 kB for a solution matching the accuracy of the seed. Similarly, on VWW, we add four new models to the Pareto frontier in the 80\% - 83\% accuracy range.
These limited improvements are primarily due to the already optimized nature of the seed models from MLPerf Tiny~\cite{mlperf-tiny}. Starting from a non-optimized DNN would make the initial SuperNet step essential prior to the application of PIT, in order to avoid utilizing sub-optimal layers.
\begin{figure}[t]
  \centering
  \includegraphics[width=0.7\columnwidth]{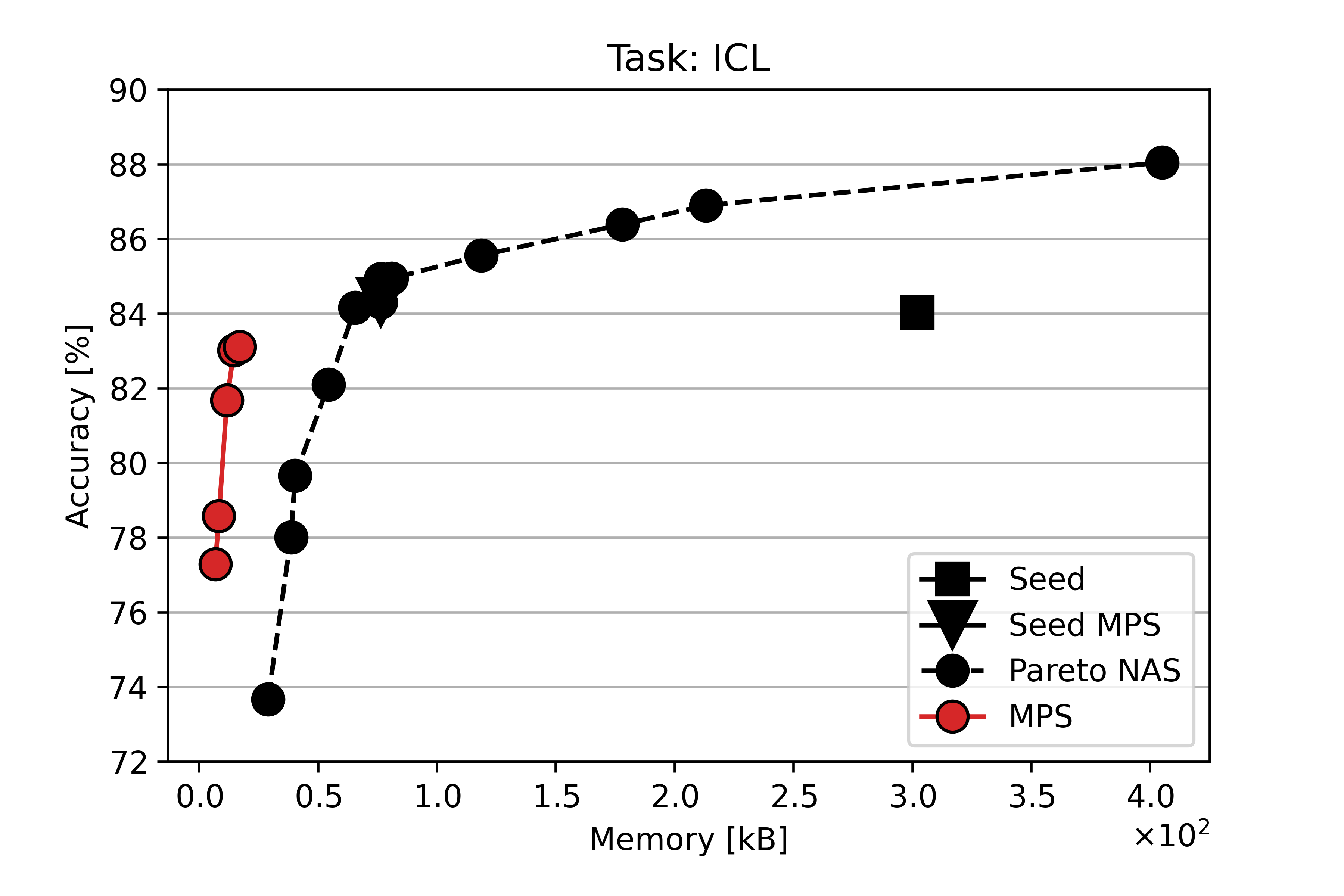}
  \vspace{-0.3cm}
  \caption{Comparison between MPS networks and floating point ones.}
  \label{fig:results_MPS}
  \vspace{-0.3cm}
\end{figure}

\subsection{Full Pipeline: SuperNet $\rightarrow$ PIT $\rightarrow$ MPS}
\rev{Fig.~\ref{fig:results_MPS} illustrates the significant improvement achieved by applying MPS on top of the DNNs obtained in the previous sections by first exporting them to standard PyTorch models and then converting them with the \texttt{plinio.MPS()} autoconversion feature.}
\rev{We search between 8-bit, 4-bit, and 2-bit integer precision and use symmetric min-max quantization and PaCT for weights and activations respectively, as in~\cite{rissoChannelwise2022}.} We show results on ICL only, due to space constraints, although MPS could be applied to all other benchmarks as well. In the graph, the dashed black line is the global Pareto curve obtained using SuperNet, PIT, or their combination. The red curve contains the new points obtained by applying MPS to the smallest architecture that matches the seed accuracy (shown as a black triangle).
Applying MPS to other Pareto-optimal architectures would be feasible too, albeit requiring more trainings. However, obtaining the complete best Pareto curve is not the focus of this work, which is rather to demonstrate the significant optimization potential unlocked by sequentially applying all PLiNIO optimizations.

The most accurate quantized architecture found with MPS reduces memory by 94.34\% compared to the seed model, with a marginal accuracy drop  of -0.92\% (83.11\% vs 84.03\%).
In comparison to the input of the MPS, it reduces memory usage by 77.66\% while sacrificing 1.82\% in accuracy. This DNN uses 8-bit quantization for all activation tensors and either 4-bit or 8-bit for weight tensors.
Additionally, by further reducing the weights/activation precision of some layers, MPS identifies several additional Pareto points.
\rev{One optimization epoch with MPS takes on average 4.3$\times$ longer compared to one reference DNN training epoch on this benchmark.}

\input{sec/12_table3}
To summarize the results, Table \ref{tab:3} shows the gains obtained applying each PLiNIO optimization (SuperNet, PIT, and MPS) in sequence on ICL, considering the smallest network that outperforms the seed model (if any) or the one achieving the highest accuracy after the optimization.

%% file: sec/10_table1.tex
\begin{table*}[]
\centering
\scriptsize
\caption{Best architectures obtained at iso-accuracy or maximizing accuracy with one between PIT or SuperNet.}
\label{tab:1}
\begin{tabular}{c|l|llll|ll}
\hline
\multicolumn{1}{l|}{Task} & Model                         & Algorithm  & Memory    & MMACs  & Accuracy & Memory-Reduction & Accuracy-improvement \\ \hline
\multirow{5}{*}{ICL}      & Seed                          & None       & 302.19 kB  & 12.5M  & 84.03 \% & n.a.             & n.a.                 \\ \cline{2-8} 
                          & \multirow{2}{*}{Iso-Accuracy} & PIT & 127.35 kB   & 7.17M  & 84.45 \% & - 56.57 \%       & + 0.42 \%            \\ \cline{3-8} 
                          &                               & SuperNet & 80.90 kB  & 5.28M  & 84.94 \% & - 73.23 \%       & + 0.91 \%            \\ \cline{2-8} 
                          & \multirow{2}{*}{Max-Accuracy} & PIT & 242.58 kB   & 9.96M  & 86.89 \% & - 19.72 \%       & + 2.86 \%            \\ \cline{3-8} 
                          &                               & SuperNet & 405.08 kB  & 18.38M & 88.05 \% & + 34.04 \%       & + 4.02 \%            \\ \hline
\multirow{5}{*}{VWW}      & Seed                          & None       & 843.33 kB & 7.49M  & 83.76 \% & n.a.             & n.a.                 \\ \cline{2-8} 
                          & \multirow{2}{*}{Iso-Accuracy} & PIT & 20.04 kB   & 1.74M  & 83.45 \% & - 97.60 \%       & - 0.31 \%            \\ \cline{3-8} 
                          &                               & SuperNet & 239.69 kB  & 11.38M & 83.28 \% & -71.27 \%        & - 0.48 \%            \\ \cline{2-8} 
                          & \multirow{2}{*}{Max-Accuracy} & PIT & 83.55 kB  & 3.3M   & 85. 88\% & - 89. 99 \%      & + 2.12 \%            \\ \cline{3-8} 
                          &                               & SuperNet & 553.52 kB  & 17.16M & 84.27 \% & - 33.66 \%       & + 0.51 \%            \\ \hline
\multirow{3}{*}{KWS}      & Seed                          & None       & 86.02 kB  & 2.66M  & 93.25 \% & n.a.             & n.a.                 \\ \cline{2-8} 
                          & \multirow{2}{*}{Max-Accuracy} & PIT & 38.91 kB   & 1.03M  & 92.82 \% & - 54.76 \%       & - 0.43 \%            \\ \cline{3-8} 
                          &                               & SuperNet & 598.79 kB & 21.0M  & 92.84 \% & + 596 \%         & - 0.41 \%            \\ \hline
\end{tabular}
\vspace{-0.4cm}
\end{table*}

%% file: sec/12_table3.tex
\begin{table}[]
\centering
\scriptsize
\caption{Optimization of a seed neural network with the three described neural architecture search applied in sequence.}
\label{tab:3}
\begin{tabular}{l|lll}
\hline
Model           & Memory & MMACs & Accuracy \\ \hline
Seed            &   302.19 kB     &  12.5M (fp32)     &    84.03\%      \\ \hline
SuperNet      &   80.90 kB (- 73.23\%)     & 5.28M (fp32)      &   84.94\%       \\ \hline
PIT      &     76.5 kB (- 74.68\%)   &   5.03M (fp32)    &   84.93\%       \\ \hline
MPS &    17.09 kB (- 94.34\%)    & 5.03M (mixed-prec)      & 83.11\%         \\ \hline
\end{tabular}
\vspace{-0.4cm}
\end{table}

%% file: sec/06_conclusions.tex
\section{Conclusions}
We presented PLiNIO, an open-source library for DNN inference optimization based on lightweight gradient-based complexity-aware techniques, including coarse- and fine-grained NAS, and MPS.
PLiNIO exposes an extendable and user-friendly interface that allows users to rapidly apply and combine these optimizations to their specific use cases.
With results on different benchmarks and DNN architectures, we have shown that PLiNIO's optimizations, combined, can generate rich Pareto-fronts in the accuracy vs memory-footprint space, \rev{reducing the size of a DNN by up to 94.34\% at almost Iso-Accuracy (-0.92\%) with respect to the baseline.}
%